\documentclass[lettersize,journal]{IEEEtran}
\usepackage{amsfonts}
\usepackage{amsmath}{
\newtheorem{assumption}{\textbf{Assumption}}
}
\usepackage{algorithm}
\usepackage{array}
\usepackage[caption=false,font=normalsize,labelfont=sf,textfont=sf]{subfig}
\usepackage{textcomp}
\usepackage{stfloats}
\usepackage{url}
\usepackage{verbatim}
\usepackage{graphicx}
\usepackage{cite}
\usepackage{xcolor}
\usepackage{cuted}
\usepackage{algpseudocode}
\usepackage{threeparttable}
\usepackage{siunitx}
\usepackage{booktabs}
\usepackage{amssymb}

\usepackage{kotex} 

\begin{document}

\title{Design, Modeling and Control of a Top-loading Fully-Actuated Cargo Transportation Multirotor}

\author{Wooyong Park, Xiangyu Wu, Dongjae Lee and Seung Jae Lee
\thanks{Wooyong Park and Seung Jae Lee are with the Department of Mechanical System Design Engineering, Seoul National University of Science and Technology (SEOULTECH), Seoul, 01811, Republic of Korea (e-mail: \{wy1004cow, seungjae\_lee\}@seoultech.ac.kr)}
\thanks{Xiangyu Wu is with Autel Robotics, Shenzhen, China (e-mail: wuxiangyu@berkeley.edu)}
\thanks{Dongjae Lee is with the Department of Aerospace Engineering, Seoul National University (SNU), and Automation and Systems Research Institute (ASRI), Seoul 08826, South Korea (e-mail: ehdwo713@snu.ac.kr)}
}



\maketitle

\begin{abstract}
Existing multirotor-based cargo transportation does not maintain a constant cargo attitude due to underactuation; however, fragile payloads may require a consistent posture.
The conventional method is also cumbersome when loading cargo, and the size of the cargo to be loaded is limited.
To overcome these issues, we propose a new fully-actuated multirotor unmanned aerial vehicle platform capable of translational motion while maintaining a constant attitude.
Our newly developed platform has a cubic exterior and can freely place cargo at any point on the flat top surface.
However, the center-of-mass (CoM) position changes when cargo is loaded, leading to undesired attitudinal motion due to unwanted torque generation.
To address this problem, we introduce a new model-free center-of-mass position estimation method inspired by the extremum-seeking control (ESC) technique.
Experimental results are presented to validate the performance of the proposed estimation method, effectively estimating the CoM position and showing satisfactory constant-attitude flight performance.
A video can be found at \url{https://youtu.be/g5yMb22a8Jo}
\end{abstract}


\section{Introduction}

 \IEEEPARstart{M}{ultirotor} unmanned aerial vehicles (UAVs) are evolving beyond simple photography/reconnaissance platforms into logistics platforms \cite{Roland,UAM}.
However, there are several issues with the flight and cargo loading methods of the current platform design.
For example, the underactuation characteristic of the flight method, where the attitude should continuously change during flight\cite{Quadrotor}, may cause cargo damage due to continuous and rapid changes in attitude during flight.
Additionally, the commonly used existing method of storing cargo in a dedicated cargo hold mounted underneath the platform makes loading/unloading difficult \cite{Cargo_grasp_1,Cargo_grasp_2}. 

To increase the logistic utility of UAVs, developing novel flight hardware that can maintain a constant attitude and conveniently load and unload cargo is necessary.
Therefore, in this study, we introduce a new fully-actuated UAV platform design, which can freely and conveniently load various volumes of cargo on the top surface of the fuselage.
We also introduce a model-free center-of-mass (CoM) position estimation method essential for the fully actuated multirotor to maintain a constant attitude while loading cargo with unknown physical properties.
Through the proposed design and controller, the platform can conveniently load/unload and transport cargo in a constant attitude throughout the flight, as if loading cargo into the compartment of a truck.

\subsection{Related works}
The keywords of this research are \textit{fully-actuated multirotor UAV} and \textit{center-of-mass position estimation}.
However, research combining these two factors is limited; therefore, we investigated related studies on each.

\subsubsection{Fully-actuated multirotor design}
Much research has been conducted focusing on developing a fully-actuated multirotor platform to enhance the applicability of multirotor UAVs \cite{Fully-actuated UAV_review}.
Platform type can be divided into two categories: fixed-tilt configurations and variable-tilt configurations.

For the fixed-tilt configurations, the thrusters are installed in fixed but various positions and directions.
Translational motion can then be controlled independently of the current attitude by controlling the magnitude and direction of the sum vector of all thrusts \cite{Fixed-tilt_1,Fixed-tilt_2,Fixed-tilt_3}.
Since the fixed-tilt configurations can create a relatively wide range of control wrenches, they can independently control translational motion while taking an extreme attitude.
However, these configurations are less suitable for cargo transportation since much energy is consumed to compensate for the internal forces among thrusters. 

In variable-tilt configurations, a servomechanism is added to control the thrust direction and enable additional control degrees of freedom (DOF).
In \cite{Variable-tilt_1,Variable-tilt_2,Variable-tilt_3}, servo actuators with one or two DOF are installed with each propeller.
However, because multi-rotor hardware has a rigid body characteristic, the minimum number of actuators required for a full DOF motion is six.
If the number of hardware actuators exceeds this, the platform becomes overactuated and redundant.
Since redundancies cause an unnecessary increase in weight and power consumption, some studies have achieved full actuation by adding only two servo mechanisms instead of four \cite{Variable-tilt_4,Variable-tilt_5}.

The variable-tilt configurations allow control of each thruster's direction independently so that almost all thrust forces can be used to overcome gravity.
This characteristic makes the variable-tilt configurations more suitable for cargo transportation than the fixed-tilt configurations.
Also, a fully-actuated platform is preferable compared to an overactuated platform since it can maintain a constant attitude during flight while minimizing weight and energy consumption.

\subsubsection{Center-of-mass position estimation}
Existing studies for estimating CoM include \cite{CoM_estimation_1,CoM_estimation_2,CoM_estimation_3}.
In \cite{CoM_estimation_1}, the maximum likelihood estimation scheme is established utilizing raw inertial measurement unit (IMU) measurements and the rotor speed data is post-processed to find the accurate CoM data.
Since the process is based on post-processing batch optimization systems, it is impossible to know the changing platform characteristics during flight.
On the contrary, in \cite{CoM_estimation_2} and \cite{CoM_estimation_3}, internal sensor-based online CoM estimation is introduced. 
However, these methods are not easily applicable in situations carrying unspecified cargo because they depend on the Kalman filter technique. 
This technique requires model information, including the physical property of the cargo, before the flight for guaranteed performance.

\subsection{Contributions}
In this study, we propose a new fully-actuated flight hardware design that can load unknown payloads freely on the flat upper area of the platform while maintaining a constant fuselage and cargo attitude during flight.
The new design places all propulsion systems inside the fuselage, giving it the additional merit of being safe from to people when flying in a populous environment.
We also propose a flight control algorithm based on the ``extremum-seeking control (ESC)" method, adaptable to changes in the physical properties of the system after loading the unknown cargo.

The remainder of the paper is structured as follows.
In Section II, we present the hardware design of the proposed platform, along with the introduction of the kinematics and dynamics of the system.
Section III introduces the controller design for full actuation of the platform, including control allocation, CoM estimation algorithms, and stability analysis.
In Section IV, we show experimental results validating the cargo transportation flight performance of the proposed system. We provide a brief summary and conclusion in Section V.

\section{Hardware}
This section introduces the design of the proposed hardware, flight principle, and the 6-DOF propulsion mechanism that brings fully actuated flight performance.
We also introduce the kinematics and dynamics of the proposed hardware.

\subsection{Hardware design}
Our hardware design aims to construct a propulsion mechanism that independently generates a three-dimensional torque vector for attitude control and a three-dimensional force vector for translational motion control.
To achieve this goal, the novel propeller tilting mechanism is configured with a minimum number of servomechanisms while keeping the physical properties of the platform during the thruster tilting motion, such as the moment of inertia (MoI).
By ensuring that the moment of inertia is invariant to the servomotor control, the entire system can be analyzed as a single rigid body, thereby securing the convenience of the controller design.

\begin{figure}[t]
\centering
\includegraphics[width=1\columnwidth]{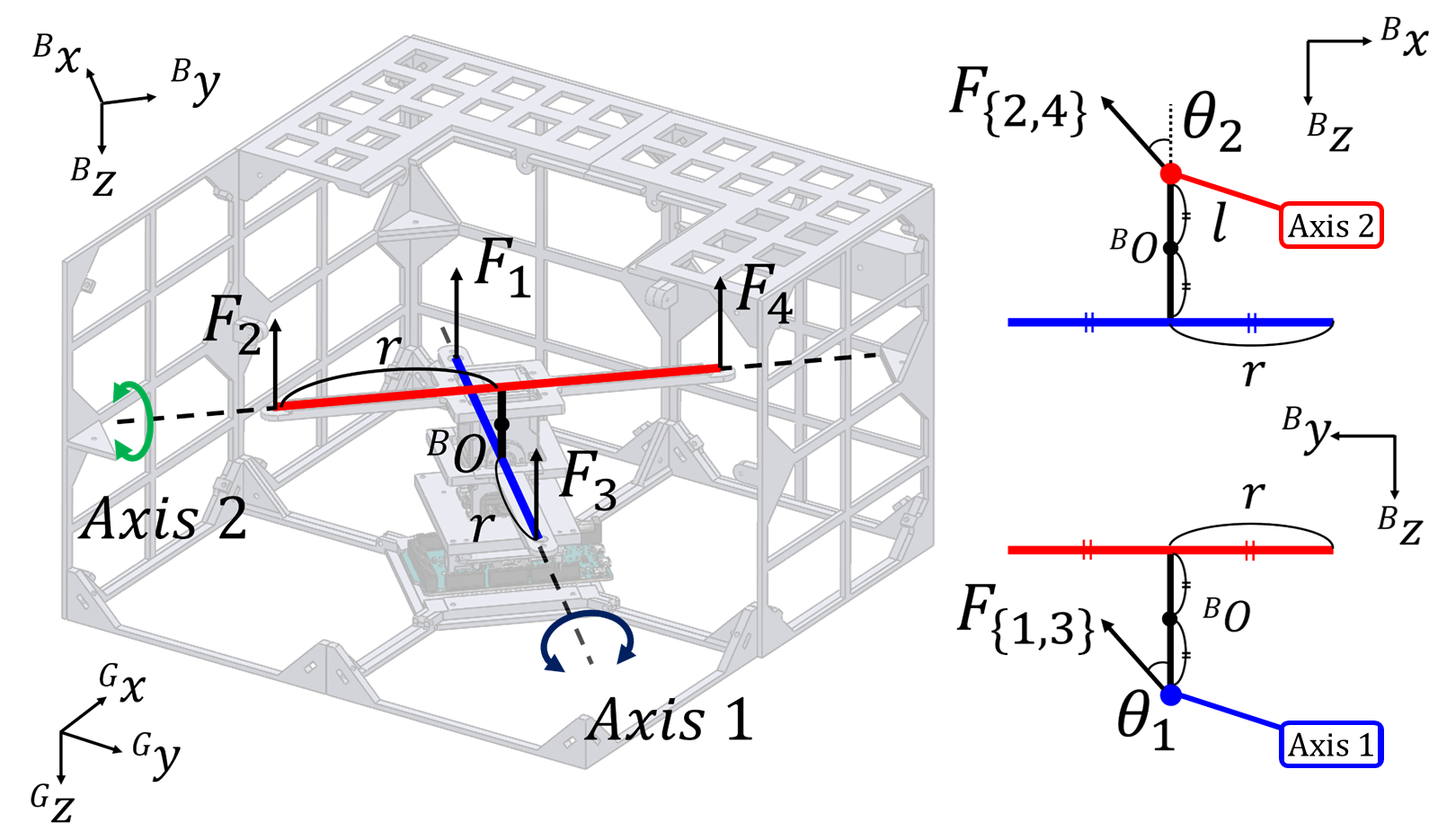}
\caption{Configuration diagram of the proposed multirotor UAV platform.}
\label{fig:Schematics}
\end{figure}

Fig. \ref{fig:Schematics} shows the structure of the proposed UAV.
The fuselage has a cubic exterior, and all propulsion systems are housed inside the fuselage.
The upper part of the fuselage has a flat surface without any protrusions, and a microstructure is installed to provide high friction between the payload and the fuselage.
This design feature allows the cargo to be placed at any point on the flat top surface of the fuselage without a dedicated cargo hold.
There are also many perforations on the top and side of the platform, minimizing changes in the aerodynamic characteristics of the propeller inside the fuselage.
The inside of the fuselage contains a column with a single Dynamixel 2XC-430 servomotor in the center, which has two servo-controlled axes perpendicular to each other, which we refer to as `Axis 1' and `Axis 2'.
Two drone arm assemblies are then positioned along each axis of the servomotor where two coaxial propeller propulsion systems are positioned at both ends of each arm.
The longitudinal principle axis of inertia for each assembly is designed to coincide with the axis of the servomotor; thus, the moment of inertia of the fuselage does not change even when the servomotor is in a rotating motion.
Through this design, thrusters 1 and 3 in Fig. \ref{fig:Schematics} can generate horizontal thrust in the body $y$ axis direction by $\theta_1$ angle control of the servomotor, and thrusters 2 and 4 can generate body $x$ axis directional force by $\theta_2$ angle control.
Ultimately, the system obtains a total of 6-DOF control utilizing two servomotors and four propulsion systems.

\subsection{Kinematics}
To realize fully actuated flight, the platform should generate three-dimensional attitude control torques and translational forces independently using the six aforementioned actuators.
Let $^B\mathbf{T}=[\tau_x\ \tau_y\ \tau_z]^T\in\mathbb{R}^{3\times1}$ and $^B\mathbf{F}=[F_x\ F_y\ F_z]^T\in\mathbb{R}^{3\times1}$ be the torque and thrust force vectors acting on the platform.
$^B(*)$ denotes that the vector is in the body-fixed frame of the platform.
We can then define a wrench vector $^B\mathbf{W}=[^B\mathbf{T}^T\ ^B\mathbf{F}^T]^T$ and manipulate it to control the pose of the platform.
A wrench vector $^B\mathbf{W}$ should be generated with a combination of six actuator outputs: $F_{\{1,2,3,4\}}$ (propeller thrusts) and $\theta_{\{1,2\}}$ (servo angles). Therefore, in this subsection, we examine the relationship between the actuator output and the wrench through a kinematics analysis of the hardware and use the information for  deriving the control allocation method in Section III.

Based on Fig. \ref{fig:Schematics}, the overall moments $^B\mathbf{T}$ generated by the set of thrusters become as follows.
\begin{equation}
    ^B\mathbf{T}=\sum_{i=1}^{4} \left(\left(^B\mathbf{r}_i-^B\mathbf{p}_c\right)_{\times}-\xi\mathbb{I}_{3\times3}\right){^B\mathbf{F}_{t,i}}
    \label{eq:torque}
\end{equation}
Here, $^B\mathbf{r}_i$ and $^B\mathbf{p}_c=[x_c\ y_c\ z_c]^T$ are position vectors of the $i$-th thruster and the CoM position of the platform, respectively, $\xi$ is a ratio between the yaw-steering reaction torque and the thrust force of the propeller thruster, $\mathbb{I}_{3\times3}$ is a $3\times3$ identity matrix, and $^B\mathbf{F}_{t,i}=[F_{t,i,x}\ F_{t,i,y}\ F_{t,i,z}]^T$ is the thrust force vector generated by the $i$-th thruster. 
$(*)_\times$ is a matrix form of the cross-product operation.

For $^B\mathbf{F}_{t,i}$, thrust force can be distributed in the horizontal and vertical directions through servomotor control.
Based on the hardware configuration shown in Fig. \ref{fig:Schematics}, the thrust vector of each motor is described as follows:
\begin{equation}
    \begin{array}{lr}
    ^B\mathbf{F}_{t,1}=[0\ \sin{\theta_1}\ -\cos{\theta_1}]^TF_1\\
    ^B\mathbf{F}_{t,2}=[-\sin{\theta_2}\ 0\ -\cos{\theta_2}]^TF_2\\
    ^B\mathbf{F}_{t,3}=[0\ \sin{\theta_1}\ -\cos{\theta_1}]^TF_3\\
    ^B\mathbf{F}_{t,4}=[-\sin{\theta_2}\ 0\ -\cos{\theta_2}]^TF_4
\end{array}.
\label{eq:force}
\end{equation}
Then, the cumulative three-dimensional force vector that controls the translational motion of the platform is as follows:
\begin{equation}
    ^B\mathbf{F}=
    \begin{bmatrix}
    -F_{A2}\sin{\theta_2}\\ 
    F_{A1}\sin{\theta_1}\\ 
    -F_{A1}\cos{\theta_1}-F_{A2}\cos{\theta_2}
    \end{bmatrix},
    \label{eq:net_forces}
\end{equation}
where $F_{A1}=F_1+F_3$ and $F_{A2}=F_2+F_4$ are collective forces generated from Axis 1 and 2 of the platform, respectively.

\subsection{Dynamics}
As mentioned above, the proposed platform has no change in the moment of inertia in any flight scenario because the principle axis of inertia of the drone arm assembly, which is the only moving part of the structure, is designed to coincide with the servo axis during the 6-DOF driving process.
Therefore, the platform can always be treated as a rigid body, and the translational and rotational motion dynamics can be modeled as follows:
\begin{equation}
\left\{
\begin{array}{lr}
    R(\mathbf{q}){^B\mathbf{F}}+m\mathbf{g}=m\ddot{\mathbf{X}}\\
    {^B\mathbf{T}}+^B\mathbf{T}_s=J{^B\dot{\mathbf{{\Omega}}}}+{^B\mathbf{\Omega}}\times J{^B\mathbf{\Omega}}
\end{array}
\right.,
    \label{eq:dynamics}
\end{equation}
where $R(\mathbf{q})\in SO(3)$ is a rotation matrix from the body coordinate to the global coordinate, $\mathbf{q}=[\phi\ \theta\ \psi]^T$ is the Euler attitude of the fuselage, and $\mathbf{g}=[0\ 0\ g]^T$ is the gravitational acceleration represented in the global frame.
$m$ is the mass, $\ddot{\mathbf{X}}\in\mathbb{R}^{3\times1}$ is the global acceleration vector, $^B\mathbf{T}_s$ is the reaction torque from the 2XC-430 servomotor which is negligible compared to the magnitude of $^B\mathbf{T}$, $J\in\mathbb{R}^{3\times3}$ is the moment of inertia tensor of the hardware, and $^B\mathbf{\Omega}=[p\ q\ r]^T\in\mathbb{R}^{3\times1}$ is the body rotation speed of the hardware. 

\begin{figure*}[b]
\vspace{-0.5cm}
\hrule
\small
\begin{equation}
    ^B\mathbf{T}=M_\tau(^B\mathbf{p}_c,\mathbf{C}_2)\mathbf{C}_1
=
    \begin{bmatrix}
    -(l-z_c)s\theta_1+y_cc\theta_1 & (r+y_c)c\theta_2+\xi s\theta_2 & -(l-z_c)s\theta_1+y_cc\theta_1 & -(r-y_c)c\theta_2+\xi s\theta_2\\
    (r-x_c)c\theta_1+\xi s\theta_1 & (l+z_c)s\theta_2-x_cc\theta_2 & -(r+x_c)c\theta_1+\xi s\theta_1 & (l+z_c)s\theta_2-x_cc\theta_2\\
    (r-x_c)s\theta_1-\xi c\theta_1 & -(r+y_c)s\theta_2+\xi c\theta_2 & -(r+x_c)s\theta_1-\xi c\theta_1 & (r-y_c)s\theta_2+\xi c\theta_2
    \end{bmatrix}
    \begin{bmatrix}
    F_1\\
    F_2\\
    F_3\\
    F_4
    \end{bmatrix}
    \label{eq:torque2}
\end{equation}
\end{figure*}

\section{Controller Design}

The operation goal of the proposed system is to fly the platform in three-dimensional space without attitudinal motion while loading arbitrary cargo randomly on the upper surface of the platform.
Therefore, we introduce a translational motion control method with a zero roll and pitch attitude in this section.

For a new platform, an unknown cargo payload causes flight control difficulties.
Since the physical properties of the loaded cargo are unknown, the values below are also unknown among the many physical characteristics of the platform:
\begin{itemize}
    \item The CoM $^B\mathbf{p}_c$ after the payload loading  is unknown.
    \item Mass $m$ is unknown.
    \item Moment of inertia $J$ is unknown.
\end{itemize}
Among these, unknown $m$ and $J$ affect translational and rotational motion control performance. Still, it is well known that stable flight is possible by using a robust controller against these uncertainties \cite{Robust_control}.
However, the unknown $^B\mathbf{p}_c$ value leads to uncertainty in wrench generation, especially the $^B\mathbf{T}$ in Equation (\ref{eq:torque}), which is the most fundamental of system control.
Therefore, to transport the unknown payload safely, estimating $^B\mathbf{p}_c$ is essential to prevent unstable and unsatisfactory flight performance.
In this section, we first introduce the control allocation method for our unique flight mechanism and then introduce dedicated model-free online $^B\mathbf{p}_c$ estimation techniques for stable 6-DOF flight control.

\subsection{Control allocation}


The new platform has two types of actuators: propeller thrusters and servomotors.
Let us define $\mathbf{C}=[\mathbf{C}_1^T\ \mathbf{C}_2^T]^T\in\mathbb{R}^{6\times1}$, where $\mathbf{C}_1=[F_1\ F_2\ F_3\ F_4]^T$ is a thruster command and $\mathbf{C}_2=[\theta_1\ \theta_2]^T$ is a servomotor command. 
We can then rewrite Equation (\ref{eq:torque}) as Equation (\ref{eq:torque2}).
Here, $M_\tau\in\mathbb{R}^{3\times4}$ is a mapping matrix between $\mathbf{C}_1$ and $^B\mathbf{T}$.
Parameters $r$ and $l$ represents arm length and servo motor dimension, respectively, as shown in Fig. \ref{fig:Schematics}.
However, $M_\tau$ includes the $\theta_{\{1,2\}}$ component, which is a component of $\mathbf{C}_2$, in a manner of multiplicative and transcendental function to other physical properties and states.
Similarly, Equation (\ref{eq:net_forces}) shows that both $\mathbf{C}_1$ and $\mathbf{C}_2$ are also used simultaneously to generate a desired $^B\mathbf{F}$ value.
Since $\mathbf{C}_1$ and $\mathbf{C}_2$ are intricate in making $^B\mathbf{W}$, we cannot allocate the actuator control input through a simple conventional mapping matrix inverse or pseudo-inverse methods \cite{Conventional_drone}.

To overcome this, we devised a sequential two-step method by utilizing the characteristics of heterogeneous actuators; the propulsion motor control response is significantly faster than the angle control response of the servomotor.

\begin{figure}[t]
\centering
\includegraphics[width=1\columnwidth]{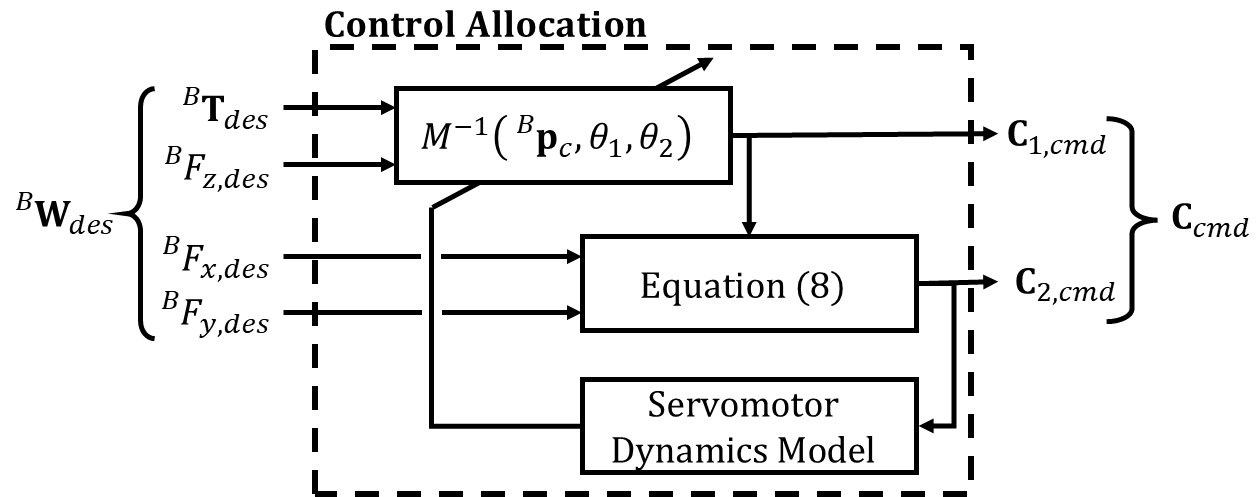}
\caption{Structure of the proposed sequential control allocation method considering heterogeneous actuator characteristics and system nonlinearity.}
\label{fig:Control_allocation}
\end{figure}

\subsubsection{Step 1}
The first step of sequential control allocation is to calculate thruster commands (i.e. $\mathbf{C}_{1,cmd}$) to achieve the desired $^B\mathbf{T}$ and $F_z$ values among the components of the wrench $^B\mathbf{W}$ in the same way as a conventional multirotor control.
From Equations (\ref{eq:net_forces}) and (\ref{eq:torque2}), the relationship between the selected states of $^B\mathbf{W}$ and $\mathbf{C}_{1,cmd}$ are as follows:
\begin{equation}
\mathbf{C}_{1,cmd}=M^{-1}(^B\mathbf{p}_c,\mathbf{C}_2)
    \begin{bmatrix}
    ^B\mathbf{T}\\
    F_z
    \end{bmatrix}_{des},
    \label{eq:C1_calculation}
\end{equation}
where
\begin{equation}
M(^B\mathbf{p}_c,\mathbf{C}_2)=
\left[
\begin{array}{cccc}
    \multicolumn{4}{c}{M_\tau(^B\mathbf{p}_c,\mathbf{C}_2)}\\
    -c\theta_1 & -c\theta_2 & -c\theta_1 & -c\theta_2 \\
\end{array} \right]\in\mathbb{R}^{4\times4}.
\label{eq:mapping}
\end{equation}
Here, $\theta_{\{1,2\}}$ constantly changes during translational motion control, which will be described in the sequel.
The $M^{-1}$ matrix requires an update of $\theta_{\{1,2\}}$ in every control iteration.
However, since the response of the propulsion motor is significantly faster than that of the servo motor, we can treat the $M$ matrix as a static map in the specific step of the control iteration.

\subsubsection{Step 2}
Once $\mathbf{C}_{1,cmd}$ is obtained through Equation (\ref{eq:C1_calculation}) in Step 1, we aquire the $F_{A\{1,2\},cmd}$ value.
Then, from Equation (\ref{eq:net_forces}), $\mathbf{C}_{2,cmd}$ can be calculated as follows:
\begin{equation}
\mathbf{C}_{2,cmd}=
    \begin{bmatrix}
    asin\left(\dfrac{F_{y,des}}{F_{A1,cmd}}\right)\\
    asin\left(\dfrac{F_{x,des}}{-F_{A2,cmd}}\right)
    \end{bmatrix}.
\end{equation}
Once the $\mathbf{C}_{2,cmd}$ is calculated, we then update the servo angles inside the $M^{-1}$ matrix of Step 1 by using the actual $\mathbf{C}_2$ signal passed through the dynamic model of the servomotor.
However, if the actual angle of the servomotor can be measured directly, this measurement can be used instead of the servomotor dynamics model.

Through the proposed sequential method, we can compute the $\mathbf{C}_{cmd}$ signal for generating $^B\mathbf{W}$. 
The overall sequential control allocation method is summarized in Fig. \ref{fig:Control_allocation}.
Next, we discuss a technique for estimating changes in the $^B\mathbf{p}_c$ value due to unspecified cargo loading.

\subsection{CoM estimation}
Let us define $^B\mathbf{p}_c=^B\hat{\mathbf{p}}_c+\Delta\mathbf{p}_c$, where $\hat{(*)}$ denotes the estimated value and $\Delta\mathbf{p}_c=[\Delta x_c\ \Delta y_c\ \Delta z_c]^T$ represents the error. 
We can then rearrange Equation (\ref{eq:torque2}) as follows:
\begin{equation}
    ^B\mathbf{T}={^B\bar{\mathbf{T}}}+{^B\Delta\mathbf{T}},\ 
    \left\{
    \begin{array}{lr}           {^B\bar{\mathbf{T}}}=M_\tau(^B\hat{\mathbf{p}}_c,\mathbf{C}_2)\mathbf{C}_1\\
        ^B\Delta\mathbf{T}={^B\mathbf{F}}_\times\Delta\mathbf{p}_c
    \end{array}
    \right.,
\label{eq:torque3}
\end{equation}
where ${^B\bar{\mathbf{T}}}$ is a nominal (or desired) torque and $^B\Delta\mathbf{T}$ is an uncertainty-driven undesired torque.
From this equation, we can see that $\Delta\mathbf{p}_c$ causes $^B\Delta\mathbf{T}$ generation during translational motion controlled by ${^B\mathbf{F}}$. 

Generation of the undesired torque is a widespread phenomenon in most UAV control, and many solutions \cite{Robust_control,drone_control_1,drone_control_2,drone_control_3}, such as integrator control of the PID controllers, high-gain control or the robust control techniques are widely utilized to compensate for static and dynamic uncertainties.
However, in our case, undesired torque is generated by $^B\mathbf{F}$, which is for translational motion control, and it changes dynamically. It also has a relatively wide bandwidth to match the needs of a high-level translational motion controller.
For the dynamic uncertainty compensation of the $^B\mathbf{T}$ signal, a robust control method such as ``Disturbance Observer (DOB)'' controller may be a tempting option; however, in the case of DOB application in actual UAV flights, the bandwidth of the disturbance that can cope with is limited due to the fuselage vibration and sensor noises. Thus, it is not suitable as a solution in our case.

Instead, we can think of estimating the accurate $^B\mathbf{p}_c$ value online during the flight after the unknown cargo is loaded; the $^B\Delta\mathbf{T}$ can then be removed systematically since $\Delta\mathbf{p}_c$ converges to zero.
For the estimation of $^B\mathbf{p}_c$, we adopted the concept of the ``ESC'' technique. 
ESC is a model-free control technique to find a local minimizer of a given time-varying cost function by applying a persistently exciting periodic perturbation to a set of chosen inputs and monitoring the corresponding output changes \cite{Xiangyu_1,Xiangyu_2}. 
The ESC concept was chosen for CoM estimation because it is model-free; the model-free method matches our system since the control allocation technique is non-linear and the physical properties of the cargo-loaded vehicle are unknown.

\begin{figure*}[t]
\centering
\includegraphics[width=\textwidth]{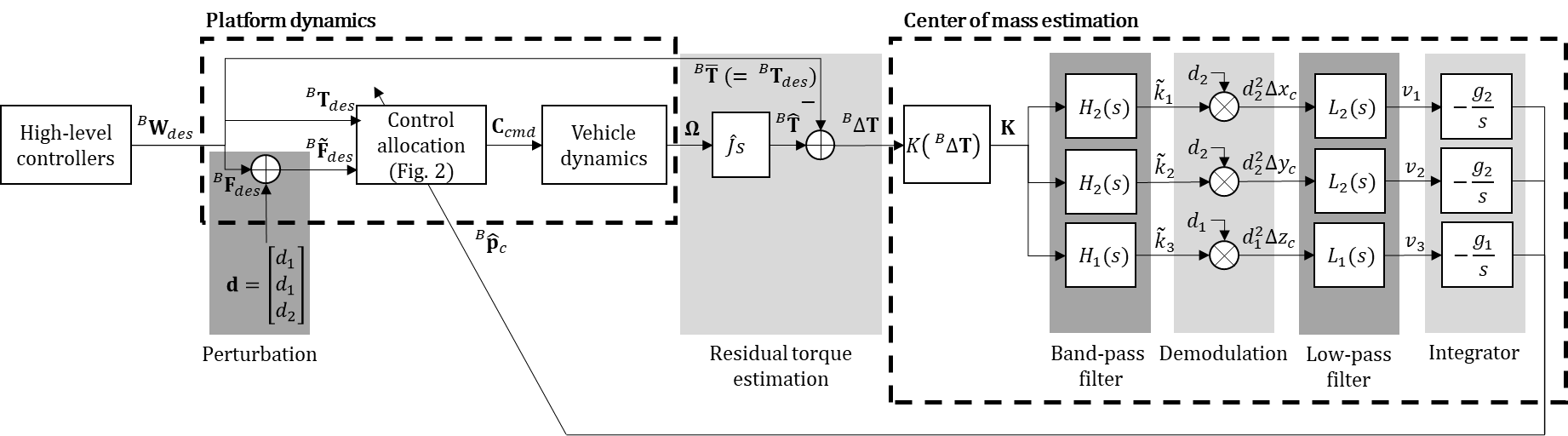}
\caption{Block diagram of the proposed flight control algorithm with online CoM estimation during flight.}
\label{fig:ESC}
\end{figure*}

The basic operation principle of the ESC algorithm is to periodically perturb some of the state variables of the platform, which is already controlled by high-level controllers, and process the perturb-induced measurements to find the gradient of a cost function to optimize the cost.
Three systems are mostly operating at the same time, and it is well known that for guaranteed system performance, the time scales must be clearly distinguishable as follows \cite{Krstic}:
\begin{itemize}
    \item Fast - plant with the (stabilizing) high-level controller,
    \item Medium - periodic perturbation,
    \item Slow - optimization algorithm.
\end{itemize}
Therefore, in our design, we also clearly distinguish the time scales of the system.

A block diagram of the proposed ESC-inspired CoM estimator is shown in Fig. \ref{fig:ESC}.
First, the control wrench $^B\mathbf{W}$ signal is generated by the ``High-level controllers'' which are the set of cascaded controllers managing the vehicle pose control.
The high-level controllers opperate the three-dimensional translational acceleration to follow target position and velocity commands while simultaneously aiming to maintain zero roll and pitch attitude.
Among the wrench signals, a dither signal $\mathbf{d}$ is added to $^B\mathbf{F}$ to get the input $^B\tilde{\mathbf{F}}$ ($={^B\mathbf{F}}+\mathbf{d}$) to the system, where $\mathbf{d}=[d_1\ d_1\ d_2]^T\in\mathbb{R}^{3\times1}$ and $d_1=a_1\sin{\omega_1t}$, $d_2=a_2\sin{\omega_2t}$.
We make the following assumptions in our stability proof of this system.
\begin{assumption}
The dither signal $\mathbf{d}$ has a relatively small amplitude and does not harm the stability of the entire system controlled by the high-level controller.
\label{assumption:dither stability}
\end{assumption}
Here, we set $\omega_1$ and $\omega_2$ far enough apart and also distinguish them from the major frequency band of the $^B\mathbf{F}$ signal.
Then, due to the $\mathbf{d}$ signal, the vehicle shows an oscillatory translation in all the $x$, $y$, and $z$ directions while simultaneously vibrating in the roll, pitch, and yaw attitudinal directions due to the $^B\Delta\mathbf{T}$ of Equation (\ref{eq:torque3}) when $^B\mathbf{p}_c$ is yet correctly estimated.

Next, we estimate the resultant attitude control torque (${^B\hat{\mathbf{T}}}$) by the ``$\hat{J}s$'' block, which is a simple differentiator with an estimated MoI tensor multiplied.
The gyroscopic effect of the airframe is small that it is permissible to neglect the term $\mathbf{\Omega}\times J\mathbf{\Omega}$ in Equation (\ref{eq:dynamics}).
The ``$\hat{J}s$ block is situated because the IMU sensor cannot directly measure rotational acceleration.
We then compare ${^B\bar{\mathbf{T}}}$ $(={^B\mathbf{T}}_{des})$ and $^B\hat{\mathbf{T}}$ to capture the $^B\Delta\mathbf{T}$ ($\approx{^B\hat{\mathbf{T}}}-{^B\bar{\mathbf{T}}}$) signal.

The full extension of the $^B\Delta\mathbf{T}$ signal in Equation (\ref{eq:torque3}) shows the following structure:
\begin{equation}
^B\Delta\mathbf{T}=
    \begin{bmatrix}
    \Delta\tau_x\\
    \Delta\tau_y\\
    \Delta\tau_z
    \end{bmatrix}
    =
    \begin{bmatrix}
    -\tilde{F}_z\Delta y_c+\tilde{F}_y\Delta z_c\\
    \tilde{F}_z\Delta x_c-\tilde{F}_x\Delta z_c\\
    -\tilde{F}_y\Delta x_c+\tilde{F}_x\Delta y_c
    \end{bmatrix}
    \label{eq:undesired_torques}.
\end{equation}
From the equation above, we can see that each element of $^B\Delta\mathbf{T}$ is a result of a combination of two translational motions in different directions, meaning that the effects of the CoM errors on two directions are indistinguishable.
A band-pass filter can be utilized to overcome this issue since the frequencies of $d_1$ and $d_2$ of $\mathbf{d}$ are set to be clearly distinguishable ($\omega_1\neq\omega_2$).
\color{red}
\color{black}

The proposed estimation process is as follows.
First, we rearrange $^B\Delta\mathbf{T}$ by a mapping function $K:\mathbb{R}^{3\times1}\rightarrow\mathbb{R}^{3\times1}$ using the following equation:
\begin{equation}
\mathbf{K}=K(^B\Delta\mathbf{T})=
\begin{bmatrix}
k_1\\
k_2\\
k_3
\end{bmatrix}
=    \begin{bmatrix}
\Delta\tau_y\\
-\Delta\tau_x\\
\Delta\tau_x\ or -\Delta\tau_y
    \end{bmatrix}.
    \label{eq:K}
\end{equation}
Then, we apply a standard second-order band-pass filter $H_{\{1,2\}}(s)$ to the $k_{\{1,2,3\}}$ signal, as shown in Fig. \ref{fig:ESC} (``Band-pass filter'' block), where
\begin{equation*}
    H_{\{1,2\}}(s)=\dfrac{\dfrac{\omega_{\{1,2\}}}{Q_{\{1,2\}}}s}{s^2+\dfrac{\omega_{\{1,2\}}}{Q_{\{1,2\}}}s+\omega_{\{1,2\}}^2}.
    \label{eq:bpfilter}
\end{equation*}
$\omega_{(*)}$ and $Q_{(*)}$ represent the natural frequency and Q-factor, respectively.
Next, we make the following assumption regarding the performance of the band-pass filter $H_{\{1,2\}}(s)$.
\begin{assumption}
If the dither signal $d_{\{1,2\}}$ is set far from the major frequency band of the $^B\mathbf{F}$ signal, and if $\Delta\mathbf{p}_c$ is updated slowly enough, $\omega_{\{1,2\}}$ and $Q_{\{1,2\}}$ exist so that the following equation holds:
\begin{equation}
\begin{array}{lr}
    \tilde{\mathbf{K}}=
    \begin{bmatrix}
    \tilde{k}_1\\
    \tilde{k}_2\\
    \tilde{k}_3
    \end{bmatrix}
    =diag(H_2,H_2,H_1)\mathbf{K}\approx
    \begin{bmatrix}
    d_2\Delta x_c\\
    d_2\Delta y_c\\
    d_1\Delta z_c
    \end{bmatrix}
    \end{array}.
    \label{eq:ktilde}
\end{equation}
\label{assumption:bandpass filter}
\end{assumption}
Here, the $H_{\{1,2\}}$ block filters out the translational motion control signal ($={}^B\mathbf{F}$) from ${}^B\tilde{\mathbf{F}}$ and outputs only  the dither signal with the specific frequency ($=d_{\{1,2\}}$), where ${}^B\tilde{\mathbf{F}}={}^B\mathbf{F}+\mathbf{d}$.
With this assumption, we can extract the $\Delta\mathbf{p}_c$ signals multiplied by the artificial dither signal among the components of the $^B\Delta\mathbf{T}$ signal in Equation (\ref{eq:undesired_torques}).
Through the demodulation process shown in Fig. \ref{fig:ESC}, we can extract the signal with the square of the dither signal for each channel ($\gamma_1$, $\gamma_2$, $\gamma_3$) where
\begin{equation}
\mathbf{\Gamma}=
\begin{bmatrix}
\gamma_1\\
\gamma_2\\
\gamma_3
\end{bmatrix}
=
\begin{bmatrix}
    d_2^2\Delta x_c\\
    d_2^2\Delta y_c\\
    d_1^2\Delta z_c
\end{bmatrix},
\end{equation}
and can find a gradient to update the $^B{\hat{\mathbf{p}}_c}$ for making $\Delta\mathbf{p}_c\rightarrow0$.

The remainder of the process is as follows.
In each channel, if the current estimate of the $\Delta\mathbf{p}_c$ element is positive, then the $\gamma_{(*)}$ signal will also be positive since the two sinusoidal signals ($\tilde{k}_{(*)}$ and demodulation signal $d_{(*)}$) are in phase.
Similarly, the $\gamma_{(*)}$ signal will be negative if the $\Delta\mathbf{p}_c$ element is negative since the two sinusoidal signals are out of phase.
In either case, the product of the two sinusoids will have a ``DC component'', which is 
 extracted by the low-pass filter to become the $\mathbf{V}=[v_1\ v_2\ v_3]^T$ signal.
Then, by integrating $\mathbf{V}$ signals with proper tunable gains $-g_{(*)}\in\mathbb{R}\leq0$ for update speed control, which must be a small gain due to the time scale separation, we can estimate the CoM values and converge $\Delta\mathbf{p}_c$ to zero \cite{Krstic} .

\subsection{Stability}
\begin{figure}[t]
\centering
\includegraphics[width=1\columnwidth]{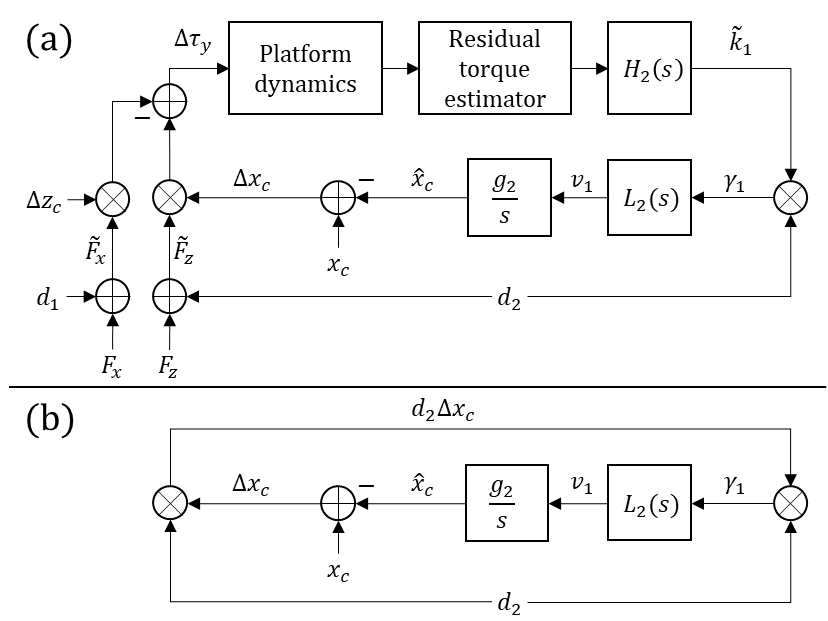}
\caption{(a) Block diagram of x-axial CoM estimation process. (b) Simplified block diagram with an application of Assumption \ref{assumption:bandpass filter}.}
\label{fig:ESC_simplified}
\end{figure}
Since the CoM estimator has the same structure for all channels, the stability of the estimator is examined by picking the $x$ channel ($x_c$ estimation).
The CoM estimation process is comprised of four steps: band-pass filter, demodulation, low-pass filter, and integrator, as shown in Fig. \ref{fig:ESC}.
However, the band-pass filtering and demodulation process must be configured to satisfy Assumptions \ref{assumption:dither stability} and \ref{assumption:bandpass filter}.
Therefore, we investigated the conditions of the low-pass filter and integrator to achieve the stability of the overall estimation process.

From Equations (\ref{eq:undesired_torques}) through (\ref{eq:ktilde}) and Fig. \ref{fig:ESC}, we can see that the update of $x_c$ is made from the $\Delta\tau_y$ signal that has passed through the $H_2(s)$ filter.
We can then simplify the $x_c$ estimation process, as shown in Fig. \ref{fig:ESC_simplified}-(a).
In addition, with the satisfactory operation of the $H_2(s)$ filter based on Assumption \ref{assumption:bandpass filter}, we can further simplify the diagram, as shown in Fig. \ref{fig:ESC_simplified}-(b).

Summarizing the system \ref{fig:ESC_simplified}-(b) gives the following results,
\begin{equation}
\frac{d}{dt}
\begin{bmatrix}
    \Delta x_c\\
    v_1
\end{bmatrix}
=
\begin{bmatrix}
    -g_2v_1\\
    -\omega_{\{L,2\}}v_1+\omega_{\{L,2\}} \gamma_1
\end{bmatrix}
\label{eq:state_space},
\end{equation}
where the low-pass filter is set to have $L_2(s)=\omega_{\{L,2\}}/(s+\omega_{\{L,2\}})$.
Our goal is to find $g_2$ and $\omega_{\{L,2\}}$ that ensure system stability in Equation (\ref{eq:state_space}).
To analyze system stability, the averaging method is utilized \cite{Krstic}.
First, let us define 
\begin{equation}
\begin{array}{lr}
\tau=\omega_2 t\\
g_2=\omega_2 \delta g_2'=O(\omega_2\delta)\\
\omega_{\{L,2\}}=\omega_2 \delta \omega_{\{L,2\}}'=O(\omega_2\delta)
\end{array},
\label{eq:condition}
\end{equation}
where $\delta$ is a small positive constant and $g_2'$ and $\omega_{\{L,2\}}'$ are $O(1)$ positive constants.
Then, Equation (\ref{eq:state_space}) becomes as follows:
\begin{equation}
    \frac{d}{d\tau}
    \begin{bmatrix}
    \Delta x_c\\
    v_1
    \end{bmatrix}=\delta
    \begin{bmatrix}
    -g_2'v_1\\
    -\omega_{\{L,2\}}'v_1+\omega_{\{L,2\}}'\Delta x_c a_2^2 \left(\sin^2(\tau)\right)
    \end{bmatrix},
    \label{eq:state_space_tau}
\end{equation}
where $\gamma_1=d_2^2 \Delta x_c$ and $d_2=a_2\sin{\omega_2 t}$.
If we set $g_2'$ to be small enough, then we can consider that  $\Delta x_c$ remains nearly constant during a single oscillation of the $d_2$ signal.
With this in mind, the average model of Equation (\ref{eq:state_space_tau}) becomes as follows:
\begin{equation}
    \frac{d}{d\tau}
    \begin{bmatrix}
    \Delta x_c^a\\
    v_1^a
    \end{bmatrix}=\delta
    \begin{bmatrix}
    -g_2'v_1^a\\
    -\omega_{\{L,2\}}'v_1^a+\dfrac{\omega_{\{L,2\}}'\Delta x_c^a a_2^2}{2\pi}\int_{0}^{2\pi}\sin^2\tau d\tau
    \end{bmatrix},
\end{equation}
which finally becomes
\begin{equation}
    \frac{d}{d\tau}
    \begin{bmatrix}
    \Delta x_c^a\\
    v_1^a
    \end{bmatrix}
    =\delta B
    \begin{bmatrix}
    \Delta x_c^a\\
    v_1^a
    \end{bmatrix},\ B=
    \begin{bmatrix}
    0 & -g_2'\\
    0.5\omega_{\{L,2\}}'a_2^2 & -\omega_{\{L,2\}}'
    \end{bmatrix},
\end{equation}
where $(*)^a$ represents the average value over a single period of oscillation.
Then, the $B$ matrix will be Hurwitz if
\begin{equation*}
    g_2',\ \omega_{\{L,2\}}'>0
\end{equation*}
and the CoM estimation process becomes stable with these inequality conditions.
However, based on the averaging analysis \cite{Khalil}, the $\delta$ value must be kept small, indicating that both $g_2$ and $\omega_{\{L,2\}}$ in Equation (\ref{eq:condition}) must also be small. 

The same principle applies to the $\Delta y_c$ and $\Delta z_c$ estimation channels, meaning that the overall conditions for the stability of online CoM estimation become
\begin{equation*}
    0<\omega_{\{L,\{1,2\}\}},\ 0<g_{\{1,2\}}
\end{equation*}
and those parameters must maintain a small value to achieve system stability.

\section{Experiment}
Experiments with two different flight scenarios were conducted to validate CoM estimation performance and the capability of stably transporting an unknown payload.
Table 1 shows the proposed vehicle's physical and control parameters.
An experiment video can be found at \url{https://youtu.be/g5yMb22a8Jo}.
\begin{table}[t]
\centering
\caption{Hardware parameters, CoM estimator parameters, and High-level controller gains}
\label{tb:physical_property}
    \begin{threeparttable}[t]
    \centering
        \begin{tabular}{lrlr}
        \toprule 
        Hardware Parameter        & Value             & Hardware Parameter          & Value     \\ 
        \midrule
        $m$               & 2.405 \si{Kg}         & $l$               & 0.015 \si{m}      \\
        $r$               & 0.109 \si{m}          & $\xi$             & 0.01     \\
        \bottomrule
        \toprule
        Estimator Parameter     & Value                     & Controller Gain    & Value    \\
        \midrule
        $a_1$, $a_2$            & 0.3, 0.7          & P (Roll, Pitch)    & 2.0    \\ 
        $\omega_1$, $\omega_2$  & 5, 3 \si{rad/s}   & D (Roll, Pitch)    & 0.45   \\
        $g_1$, $g_2$            & 1.5, 0.5          & P (X, Y Position)  & 2.0    \\
        $Q_{\{1,2\}}$           & 20                & I (X, Y Position)  & 0.5     \\
        $\omega_{\{L,\{1,2\}\}}$& 0.5 \si{rad/s}    & D (X, Y Position)  & 0.7     \\
        \bottomrule
        \end{tabular}
    \end{threeparttable}
    \label{table:parameters}
\end{table}



\subsection{CoM estimation performance}

\begin{figure*}[t]
\centering
\includegraphics[width=1.97\columnwidth]{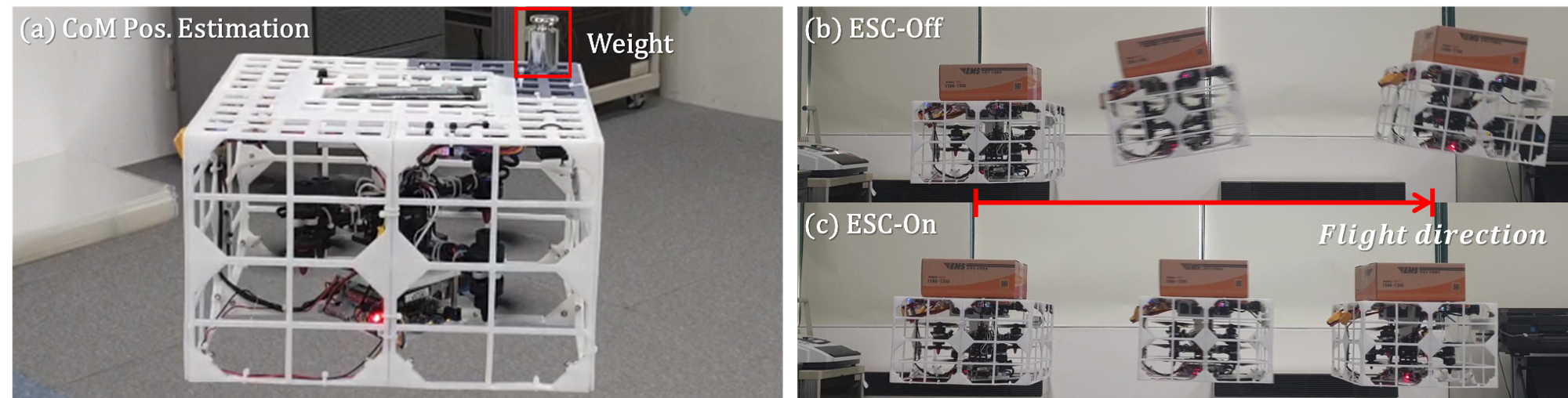}
\caption{Experiment environment for performance validation. (a) Experiment to estimate the position of the changed CoM after attaching a weight to a fixed position. (b) Flight experiment in which a constant attitude is not maintained during translational motion due to ill-estimated CoM when the proposed ESC is turned off. (c) Flight experiment maintaining a constant attitude during translational motion with updated CoM with the proposed ESC enabled.}
\label{fig:Experimental_settings}
\end{figure*}

\begin{figure*}[t]
\centering
\includegraphics[width=2\columnwidth]{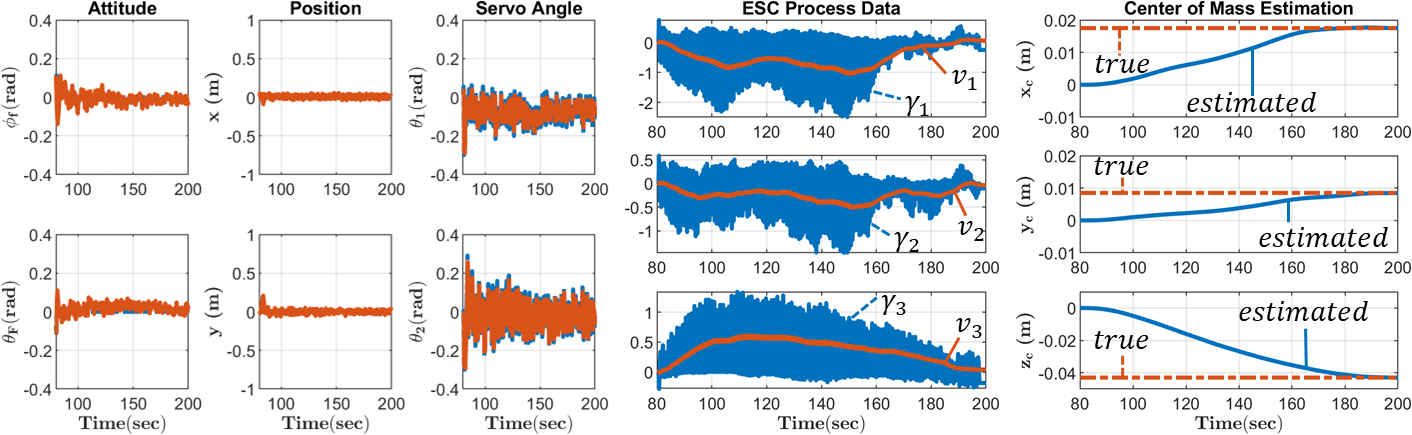}
\caption{Flight data of the proposed CoM localization algorithm. (Left three columns) The states of the platform vibrate due to the dither signal. The blue dotted line is the desired value, and the solid orange line is the sensor data. (Middle column) $\mathbf{\Gamma}$ and $\mathbf{V}$ vector signals that change over time during the CoM estimation process. (Right column) Vector of CoM estimates that change over time. The dotted orange line is the true CoM. } 
\label{fig:CoM_Estimation}
\end{figure*}

The first experiment was conducted to validate the performance of the CoM estimation algorithm. 
As shown in Fig. \ref{fig:Experimental_settings}-(a), in this scenario, the weight with a known mass (0.2 \si{Kg}) is attached at a specific position ($[0.184 \ 0 \ -0.121]^T\ \si{m}$ in the body frame); thus, the actual CoM of the entire system is known ($^B{\mathbf{p}}_c=[0.0175\ 0.0085\ -0.0430]^T \si{m}$), derived from the CAD design tool. 
The CoM estimation algorithm is then activated to validate the performance of the algorithm.

Fig. \ref{fig:CoM_Estimation} shows the flight result. 
The left three columns of the figure show fuselage's roll and pitch attitude, horizontal positions, and two servo angles, respectively. 
The high-level controller attempts to maintain zero roll and pitch attitude and zero $x$ and $y$ positions. 
However, due to the dither signal $\mathbf{d}$, an undesired attitude-control torque $^B{\Delta\mathbf{T}}$ is generated, and the attitude oscillates.
Here, we can see that the position of the platform remains at zero due to the robustness of the high-level position controller. 

The middle and right columns of the figure show the internal process of the CoM estimation algorithm. 
In the middle column graphs (``ESC Process Data'' graphs), the blue signals are $\mathbf{\Gamma}$ signals, and the orange signals are $\mathbf{V}$ signals, which are low-pass filtered signals.
The $\mathbf{V}$ signal acts as a gradient of the estimator, where the negative and positive values indicate that the estimated CoM must increase or decrease. 
In the end, as we see in the graphs on the right column (``Center of Mass Estimation'' graphs), the estimation algorithm successfully estimates the actual CoM. 

\subsection{Flight experiment carrying an unknown payload}
The second experiment is actual payload transportation. A payload with unknown physical properties is loaded on the platform. 
Two flight scenarios are conducted to validate the performance of the proposed algorithm.
In both scenarios, a position control command is applied making the platform reciprocate about three meters in the $y$-direction.
Also, the desired roll and pitch attitude of the high-level are set to be zero in both scenarios.

\subsubsection{Scenario 1}
\begin{figure}[t]
\centering
\includegraphics[width=1\columnwidth]{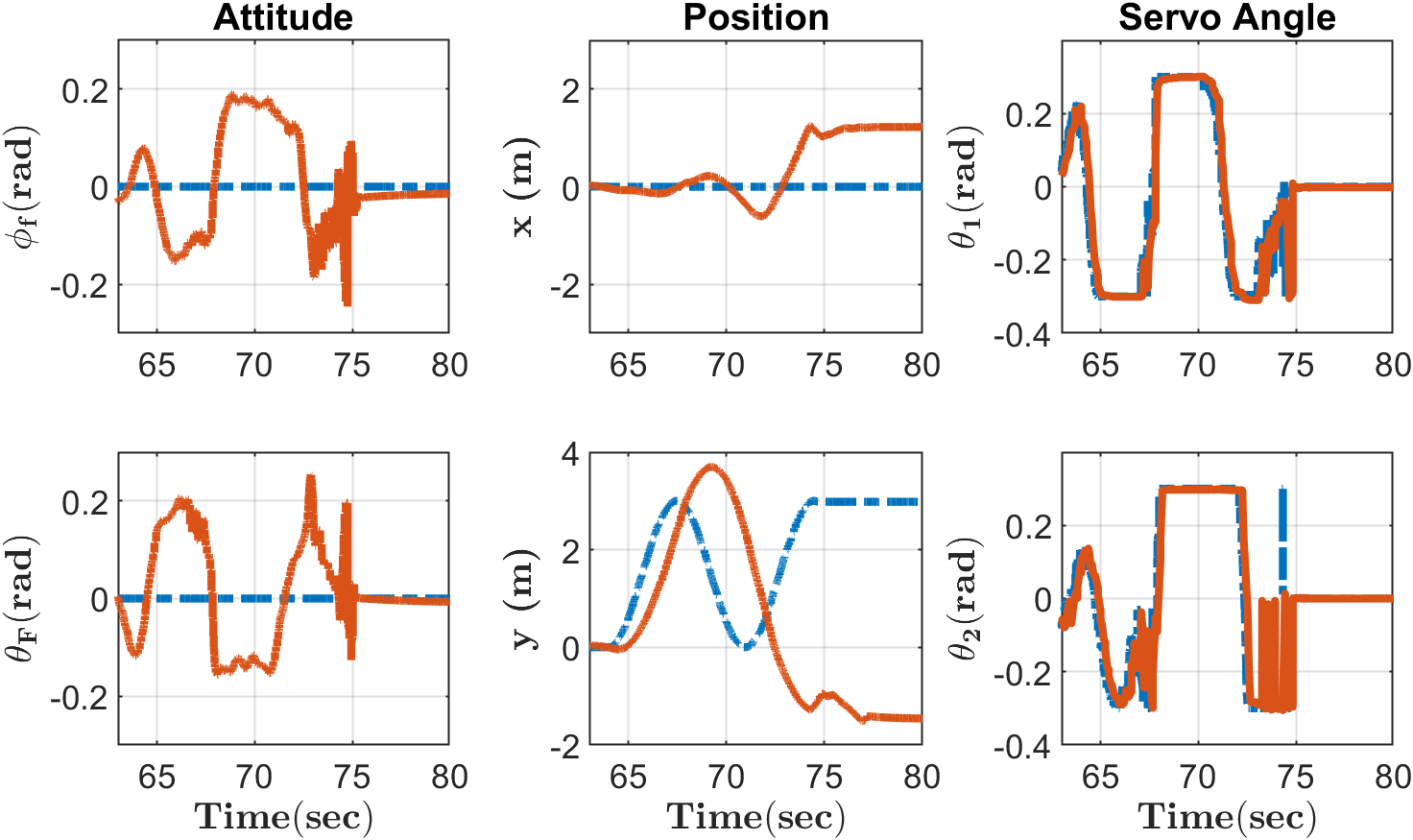}\centering
\caption{Cargo flight results without CoM estimation (blue dashed line: desired value, orange solid line: sensor data). The fuselage attitude is not maintained due to undesired torque generation; as a result, the servo angle reaches the hardware limit, and the overall control fails.}
\label{fig:Experiment_ESC_off}
\end{figure}

Fig. \ref{fig:Experiment_ESC_off} shows the flight without CoM estimation.
From around 65 seconds, the $y$-directional translational motion begins. At this time, undesired roll and pitch motions occur due to undesired torque.
The sequential image of the experiment is shown in Fig. 5-(b). 

Failure to maintain the attitude of the platform results in severe impairment of translational motion control.
As can be seen in the ``Servo Angle'' graphs, the rotation of each servomotor is limited to $\pm0.3$ \si{rad} due to hardware limitations. Because of this, the platform can no longer generate the necessary global horizontal thrust force since the fuselage has been severely tilted.
Ultimately, the platform collided with the environment at approximately 75 seconds and crashed.

\subsubsection{Scenario 2}
\begin{figure}[t]
\centering
\includegraphics[width=1\columnwidth]{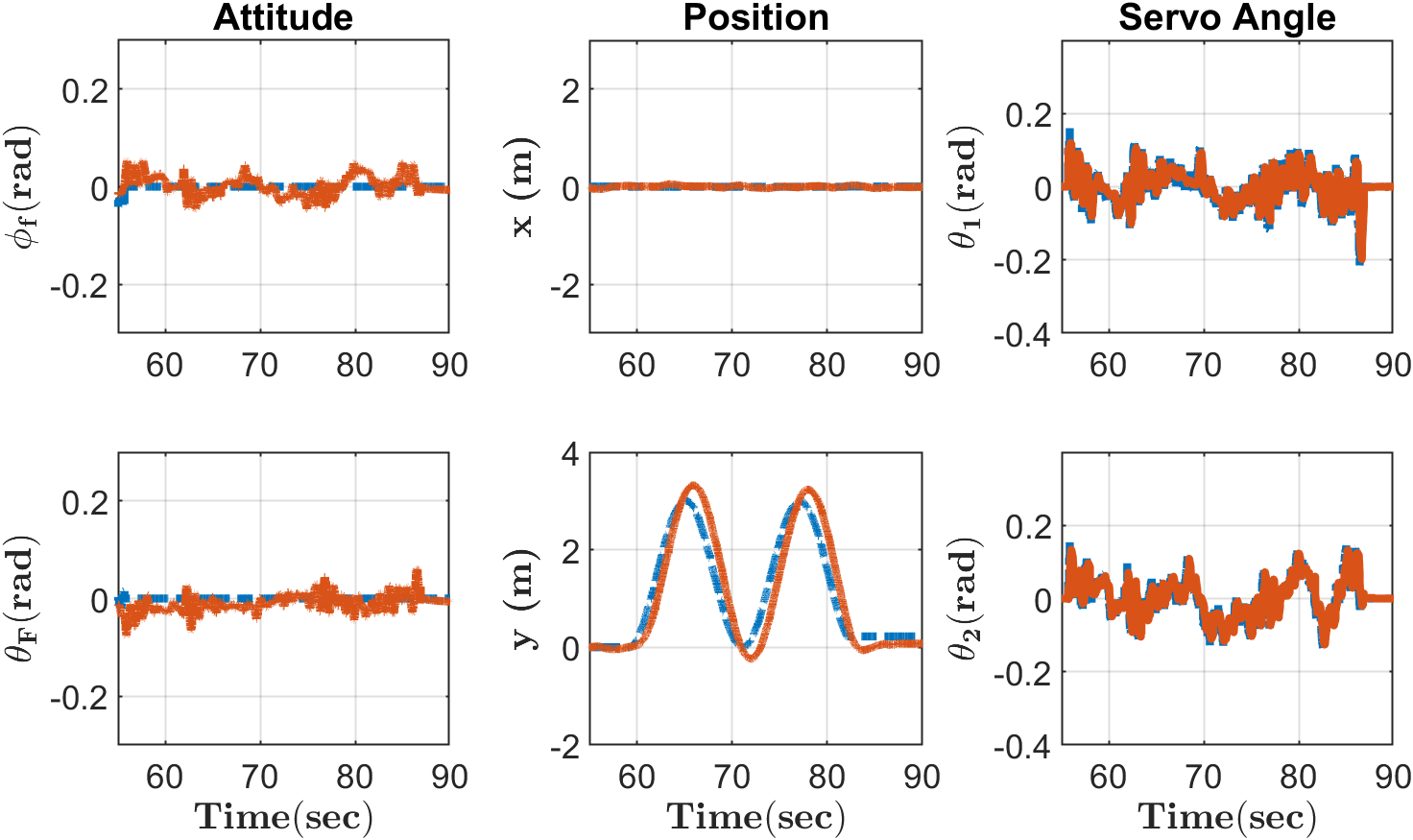}\centering
\caption{Cargo flight results with CoM estimation (blue dashed line: desired value, orange solid line: sensor data). The fuselage attitude is maintained within $\pm0.05$ \si{rad} ($\pm2.86$ \si{deg}), and the position control is also successful.}
\label{fig:Experiment_ESC_on}
\vspace{-0.3cm}
\end{figure}

Conversely, Fig. \ref{fig:Experiment_ESC_on} shows the flight with the estimated CoM.
Since the CoM is updated, the undesired torque generation is minimized; therefore, the platform can maintain near-zero roll and pitch attitude during the flight.
The translational motion performance is also satisfactory due to the thrust control by the two servo motors.
The sequential image of the experiment is shown in Fig. 5-(c). 

\section{Conclusion}
In this research, we introduced a new multirotor UAV platform and a dedicated control method suitable for unknown cargo transportation.
A platform with a new fully-actuated flight mechanism that can pursue stable cargo transport while maintaining a constant attitude was developed to achieve this goal.
The platform has a cubic-shaped exterior designed to freely place unknown cargo anywhere on the flat upper space.
A model-free CoM estimation technique inspired by the ESC algorithm was introduced to overcome the deterioration of the attitude control performance due to undesired torque caused by unknown cargo.
For estimation, dither signals having different frequencies are applied to the three-dimensional translational force signal by utilizing the 6-DOF flight performance of the fully-actuated platform. 
An accurate CoM is then estimated by monitoring the attitudinal vibration.
During the estimation process, each roll, pitch, and yaw attitude vibration is caused by three translational dither forces, and a band-pass filter is introduced to distinguish the effect of each dither force in each attitude channel.
Finally, the estimation and flight performance of the CoM was validated through experiments.

The current study is limited in that the estimation process was performed before the translational motion.
The requirement of a separate time period for estimation can be a weakness in battery-based aircraft with a limited flight time.
Therefore, future research will focus on a technique that can quickly estimate the physical properties while in motion.






\vfill


\begin{thebibliography}{1}
\bibliographystyle{IEEEtran}

\bibitem{Roland}
Baur, Stephan, et al. ``Cargo drones: the future of parcel delivery." Roland Berger, 19 Feb. 2020, \url{https://www.rolandberger.com/en/Insights/Publications/Cargo-drones-The-future-of-parcel-delivery.html/}.

\bibitem{UAM}
Straubinger, Anna, et al. ''An overview of current research and developments in urban air mobility–Setting the scene for UAM introduction." Journal of Air Transport Management 87 (2020): 101852.

\bibitem{Quadrotor}
Emran, Bara J., et al. ''A review of quadrotor: An underactuated mechanical system." Annual Reviews in Control 46 (2018): 165-180.

\bibitem{Cargo_grasp_1}
Lindsey, Quentin, et al. ''Construction of cubic structures with quadrotor teams." Proc. Robotics: Science \& Systems VII 7 (2011).

\bibitem{Cargo_grasp_2}
Xiang, Gang, et al. ''Design of the life-ring drone delivery system for rip current rescue." 2016 IEEE Systems and Information Engineering Design Symposium. IEEE, 2016.


\bibitem{Fully-actuated UAV_review}
Rashad, Ramy, et al. ''Fully actuated multirotor UAVs: A literature review." IEEE Robotics \& Automation Magazine 27.3 (2020): 97-107.

\bibitem{Fixed-tilt_1}
von Frankenberg, et al. ''Disturbance rejection in multi-rotor unmanned aerial vehicles using a novel rotor geometry." Proc. 4th Int. Conf. Control Dynamic Systems and Robotics. 2017.

\bibitem{Fixed-tilt_2}
Brescianini, Dario, et al. ''Design, modeling and control of an omni-directional aerial vehicle." 2016 IEEE international conference on robotics and automation. IEEE, 2016.

\bibitem{Fixed-tilt_3}
Nikou, Alexandros, et al. ''Mechanical design, modelling and control of a novel aerial manipulator." 2015 IEEE International Conference on Robotics and Automation. IEEE, 2015.

\bibitem{Variable-tilt_1}
Badr, Sherif, et al. ''A novel modification for a quadrotor design." 2016 International Conference on Unmanned Aircraft Systems. IEEE, 2016.

\bibitem{Variable-tilt_2}
Ryll, Markus, et al. ''A novel overactuated quadrotor unmanned aerial vehicle: Modeling, control, and experimental validation." IEEE Transactions on Control Systems Technology 23.2 (2014): 540-556.

\bibitem{Variable-tilt_3}
Kamel, Mina, et al. ''The voliro omniorientational hexacopter: An agile and maneuverable tiltable-rotor aerial vehicle." IEEE Robotics \& Automation Magazine 25.4 (2018): 34-44.

\bibitem{Variable-tilt_4}
Segui-Gasco, Pau, et al. ''A novel actuation concept for a multi rotor UAV." Journal of Intelligent \& Robotic Systems 74.1 (2014): 173-191.

\bibitem{Variable-tilt_5}
Lee, Seung Jae, et al. ''Fully actuated autonomous flight of thruster-tilting multirotor." IEEE/ASME Transactions on Mechatronics 26.2 (2020): 765-776.

\bibitem{CoM_estimation_1}
Burri, Michael, et al. ''A framework for maximum likelihood parameter identification applied on MAVs." Journal of Field Robotics 35.1 (2018): 5-22.

\bibitem{CoM_estimation_2}
Wüest, Valentin, et al. ''Online estimation of geometric and inertia parameters for multirotor aerial vehicles." 2019 International Conference on Robotics and Automation. IEEE, 2019.

\bibitem{CoM_estimation_3}
Lee, Hyeonbeom, et al. ''Estimation and Control of Cooperative Aerial Manipulators for a Payload with an Arbitrary Center-of-Mass." Sensors 21.19 (2021): 6452.

\bibitem{Robust_control}
Kim, Suseong, et al. ''Robust control of an equipment-added multirotor using disturbance observer." IEEE Transactions on Control Systems Technology 26.4 (2017): 1524-1531.

\bibitem{Conventional_drone}
Pounds, Paul Edward Ian. ''Design, construction and control of a large quadrotor micro air vehicle." (2007).

\bibitem{drone_control_1}
Kada, Belkacem, et al. ''Robust PID controller design for an UAV flight control system." Proceedings of the World congress on Engineering and Computer Science. Vol. 2. No. 1-6. 2011.

\bibitem{drone_control_2}
Santoso, Fendy, et al. ''Robust hybrid nonlinear control systems for the dynamics of a quadcopter drone." IEEE Transactions on Systems, Man, and Cybernetics: Systems 50.8 (2018): 3059-3071.

\bibitem{drone_control_3}
Lee, Seung Jae, et al. ''Robust translational force control of multi-rotor uav for precise acceleration tracking." IEEE Transactions on Automation Science and Engineering 17.2 (2019): 562-573.

\bibitem{Xiangyu_1}
Tagliabue, Andrea, et al. ''Model-free online motion adaptation for optimal range and endurance of multicopters." 2019 International Conference on Robotics and Automation. IEEE, 2019.

\bibitem{Xiangyu_2}
Wu, Xiangyu, et al. ''Model-free online motion adaptation for energy-efficient flight of multicopters." IEEE Access 10 (2022): 65507-65519.

\bibitem{Krstic}
Krstic, Miroslav, et al. ''Stability of extremum seeking feedback for general nonlinear dynamic systems." Automatica-Kidlington 36.4 (2000): 595-602.

\bibitem{Khalil}
Khalil, Hassan K. Nonlinear control. Vol. 406. New York: Pearson, 2015.




\end{thebibliography}
\end{document}